\documentclass[letterpaper]{article} 
\usepackage[preprint]{aaai2027}  
\usepackage[hyphens]{url}  
\usepackage{graphicx} 
\usepackage{natbib}  
\usepackage{caption} 
\usepackage{algorithm}
\usepackage{algorithmic}
\usepackage{booktabs}
\usepackage{amsmath}
\usepackage{xspace}
\usepackage{soul}
\usepackage{multirow}

\usepackage{adjustbox}
\usepackage[table]{xcolor}
\usepackage{float}

\newcommand{\methodname}{GyroNovo\xspace}

\title{\methodname: Error-Guided Fragment Imputation with Mass-Aware Attention for \textit{De Novo} Peptide Sequencing}

\author{
    Abdellah El Mekki\textsuperscript{\rm 1},
    Laks V.S. Lakshmanan\textsuperscript{\rm 1},
    Muhammad Abdul-Mageed\textsuperscript{\rm 1,2}
}
\affiliations{
    \textsuperscript{\rm 1}The University of British Columbia\\
    \textsuperscript{\rm 2}Canada Research Chair in NLP and ML\\
    \{abdellah.elmekki, muhammad.mageed\}@ubc.ca, laks@cs.ubc.ca
}

\begin{document}

\maketitle

\begin{abstract}

\textit{De novo} peptide sequencing from tandem mass spectra is essential for identifying peptides without relying on reference databases. Despite recent advances in deep learning, accurate sequencing remains challenging because experimental spectra are often sparse, noisy, and incomplete, causing informative b- and y-ion fragments to be unobserved. Existing methods have attempted to recover this missing evidence through latent-space imputation before autoregressive decoding. However, they typically formulate imputation as a fixed reconstruction task, without considering which missing fragments are most relevant to the decoder’s current sequencing errors. Moreover, existing peak representations do not explicitly model mass differences between peaks, despite their fundamental importance in \textit{de novo} peptide sequencing.

We introduce \methodname, a framework with two main contributions. First, we use decoder errors observed during training to adapt the imputation objective, assigning greater emphasis to fragments associated with frequent decoding errors. We further use the decoder error distribution to construct easy and hard augmented views of each peptide spectrum, enabling the decoder to learn under varying degrees of spectral corruption and missing-fragment severity. Second, we introduce a mass-aware inductive bias into self-attention by using rotary embeddings to encode pairwise mass differences between spectral peaks. Together, these components align missing-fragment recovery with the decoder’s sequencing behavior while explicitly incorporating the mass relationships that underlie peptide fragmentation. At inference time, \methodname retains a standard encoder–imputer–decoder architecture and requires neither additional inputs nor auxiliary search procedures. Experiments on NovoBench demonstrate improvements of approximately 9 percentage points in peptide-level precision and 7 percentage points in amino-acid-level precision over the previous state-of-the-art baseline. Our code is publicly available at \url{https://github.com/UBC-NLP/gyronovo}.

\end{abstract}


\section{Introduction}

\textit{De novo} peptide sequencing is a central task in computational proteomics: given a tandem mass spectrum, the goal is to infer the underlying amino-acid sequence without relying on a protein database. Accurate \textit{de novo} sequencing enables discovery-oriented workflows, including antibody and antigen characterization \cite{Tran2019}, supports the analysis of organisms or samples with incomplete or unavailable reference databases \cite{Yilmaz2024}, and helps identify novel or modified peptides that are difficult to recover using conventional database search \cite{doi:10.1142/S0219720013500078,Chick2015}. Recent neural approaches, including influential Transformer-based sequence decoders \cite{Yilmaz2024} and other systems systematically evaluated in NovoBench \cite{zhou2024novobench}, have substantially advanced the field by learning rich representations of mass spectra and decoding peptide sequences directly from observed peaks. Despite these advances, practical spectra remain difficult because fragmentation is incomplete, ion intensities are irregular, noise peaks are common, and post-translational modifications (PTMs) can obscure otherwise useful mass evidence \cite{doi:10.1016/j.csbj.2022.03.008}.

A major source of sequencing error is missing fragmentation evidence \cite{Mao2023, du2025latent}. Autoregressive decoders predict each amino acid from a partial sequence and a spectrum representation, even when evidence for the relevant cleavage is weak or absent. LIPNovo addresses this limitation through latent imputation: learnable peak queries infer representations of theoretical fragments, and the peptide decoder jointly uses observed and imputed evidence \cite{du2025latent}. This design directly targets missing fragments and provides a strong foundation for robust \textit{de novo} sequencing. However, LIPNovo does not explicitly prioritize the missing fragments most useful for correcting current decoding errors. Its imputation supervision is determined mainly by theoretical peak targets and bipartite matching, which encourages plausible fragment reconstruction but treats easy and difficult residue decisions similarly. As a result, the model may emphasize valid peaks with limited downstream value while underweighting fragments whose absence causes sequencing errors.

Current methods also provide minor explicit structure for relationships among observed peaks. Tandem mass spectra are organized by continuous mass differences rather than discrete sequence positions: residue transitions, fragmentation ladders, complementary ions, and neutral losses all induce characteristic mass separations. Standard transformer encoders represent each peak using absolute m/z and intensity, leaving self-attention to infer these relative relationships implicitly. Introducing a relative-mass inductive bias could therefore better align the encoder with the geometry of tandem mass spectra.

We propose \methodname, a method that addresses both limitations through decoder-guided latent imputation and continuous mass-aware rotary attention. During training, token-level decoder errors are mapped to nearby theoretical fragment cleavages, assigning greater imputation weight to evidence associated with difficult residue decisions. The same feedback guides peptide-conditioned spectral augmentation, emphasizing informative spectra and cleavage-local regions while constructing easier and harder views that preserve observed peaks and selectively modify theoretical fragments. We also introduce continuous mass-based rotary attention in the spectrum encoder. Instead of using peak-list indices, rotation angles are derived from each peak’s continuous m/z coordinate at multiple wavelengths. Attention therefore depends on relative mass displacement, enabling multi-resolution modeling of mass relationships.

In summary, this work makes the following contributions: (i) a decoder-conditioned latent imputation objective that projects token-level sequencing errors onto cleavage-local theoretical fragment targets; (ii) an observed-anchor, peptide-conditioned strategy that produces realistic easy and hard spectral views with both imputation and decoding supervision; (iii) a continuous mass-based rotary self-attention mechanism that introduces relative mass differences directly into the spectrum encoder; and (iv) an empirical NovoBench evaluation demonstrating state-of-the-art \textit{de novo} peptide sequencing, with approximately 9- and 7-percentage-point gains in peptide- and amino-acid-level precision, respectively.

\section{Related Work}

\subsection{\textit{De Novo} Peptide Sequencing}
Classical \textit{de novo} sequencing methods formulate peptide identification as path finding or dynamic programming over spectrum graphs, using handcrafted or probabilistic scoring functions \cite{https://doi.org/10.1002/rcm.1196,10.1021/ac048788h, Ma2015Novor}. DeepNovo introduced neural next-residue prediction, followed by architectures that model spectra as unordered peak sets or graphs, including PointNovo, GraphNovo, and Spectralis \cite{doi:10.1073/pnas.1705691114,10.1093/bioadv/vbad057,Mao2023,Klaproth-Andrade2024}. Nevertheless, sequencing accuracy remains highly sensitive to noise and missing fragment ions \cite{zhou2024novobench}.

Deep learning models for mass spectrometry have employed convolutional, recurrent, point-set, and graph architectures to learn representations directly from irregular MS/MS peaks. Most methods encode the observed spectrum and immediately decode the peptide. LIPNovo instead imputes latent representations of theoretical fragment peaks before prediction, directly targeting incomplete fragmentation \cite{du2025latent}.

\subsection{Transformer-based \textit{De Novo} Peptide Sequencing}

Transformers provide a natural encoder-decoder formulation for translating spectra into peptide sequences \cite{3295222.3295349}. Casanovo introduced direct autoregressive decoding from peak sequences, while subsequent methods such as ContraNovo, InstaNovo, $\pi$-PrimeNovo, AdaNovo, and $\pi$-HelixNovo improved scalability, robustness, or decoding \cite{Yilmaz2024, jin2024contranovo,Eloff2025, Zhang2025, xia2024adanovo, 10.1093/bib/bbae021}. Conventional peak embeddings primarily encode absolute mass-to-charge values, requiring chemically meaningful mass differences to be learned implicitly. Recent methods incorporate pairwise mass differences through attention biases \cite{10.1021/acs.jproteome.5c00063}.

\section{Proposed Method}

\begin{figure*}[h]
    \centering
    \includegraphics[width=0.99\textwidth]{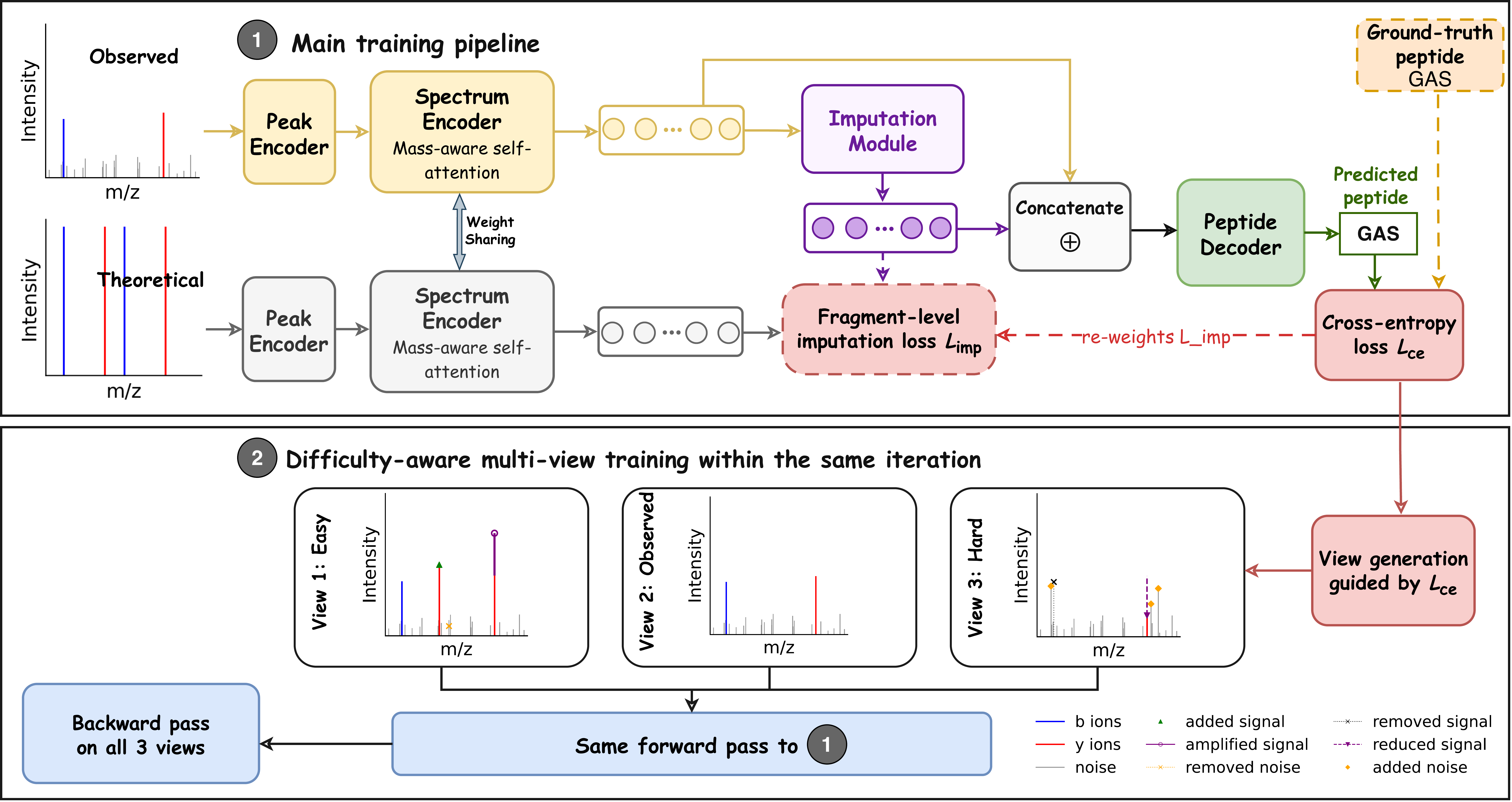}
    \caption{Overview of \methodname’s error-guided imputation and difficulty-aware multi-view training framework.}
    \label{fig:main_fig}
\end{figure*}

\subsection{Problem Formulation}

We consider the problem of \textit{de novo} peptide sequencing, whose objective is to recover the amino acid sequence of a peptide from an observed tandem mass spectrum and its precursor information. An MS/MS spectrum is represented as a variable-size set of peaks
$
x=\{(m_i,I_i)\}_{i=1}^{N},
$
where \(m_i\) and \(I_i\) denote the mass-to-charge ratio and intensity of the \(i\)-th peak, respectively. Each spectrum is accompanied by precursor metadata \(p=(M,z)\), where \(M\) is the precursor mass and \(z\) is the precursor charge. The target peptide is represented as
$
y=(y_1,\ldots,y_L),
$
where $y_l$ is an amino acid from a vocabulary containing both unmodified and modified residues, together with a stop token. The sequencing model factorizes the conditional probability of the peptide autoregressively:
\[
P(y\mid x,p)
=
\prod_{l=1}^{L}
P(y_l\mid y_{<l},x,p).
\]
Unlike database-search methods, \textit{de novo} sequencing performs this prediction without access to a protein sequence database.

A central challenge is that experimental spectra provide only a partial and noisy observation of the underlying fragmentation process. Informative \(b\)- and \(y\)-ions may be absent, have low intensity, or overlap with unrelated noise peaks. Consequently, the decoder must often infer amino acids from incomplete fragment-ion ladders.

A common neural formulation, exemplified by Casanovo~\cite{Yilmaz2024}, uses a Transformer encoder to represent the observed spectrum and an autoregressive Transformer decoder to generate the peptide sequence. LIPNovo~\cite{du2025latent} extends this formulation with a latent imputation module that predicts representations of missing fragment ions before decoding. In this work, we retain the latent imputation backbone of LIPNovo and introduce \methodname which has two key innovations: (i) decoder-conditioned supervision that concentrates imputation and augmentation on fragments associated with difficult amino acids, and (ii) continuous mass rotary attention that makes spectrum self-attention directly sensitive to relative mass differences.

\subsection{Latent Imputation Backbone}

We follow LIPNovo~\cite{du2025latent} for the latent imputation backbone. Each observed peak is first mapped to a peak embedding by combining a continuous encoding of its m/z value with a learned projection of its intensity. A spectrum encoder \(E_\theta\) then produces contextualized peak representations:
\[
H
=
E_\theta(x,p)
=
[h_{\mathrm{cls}},h_1,\ldots,h_N],
\]
where \(h_{\mathrm{cls}}\) is a learned global spectrum representation and \(h_i\) is the contextual representation of the \(i\)-th peak.

An imputation module \(G_\theta\) uses \(Q\) learned queries to cross-attend to the encoded spectrum:
\[
\{(\hat{u}_j,\hat{r}_j)\}_{j=1}^{Q}
=
G_\theta(H).
\]
Here, \(\hat{u}_j\) is a candidate latent fragment representation and \(\hat{r}_j\) is its confidence logit.

During training, the ground-truth peptide is used to construct a theoretical spectrum containing its \(b\)- and \(y\)-ions. The theoretical spectrum is processed by the same spectrum encoder, yielding target fragment representations
\[
U^\star=\{u_k^\star\}_{k=1}^{K}.
\]
The predicted queries and theoretical fragments are aligned through bipartite matching. For a training sample \(i\), the matching is obtained by minimizing
\[
\pi_i
=
\arg\min_{\pi}
\sum_{(j,k)\in\pi}
\left[
\|\hat{u}_{ij}-u_{ik}^\star\|_2^2
+
1-\sigma(\hat{r}_{ij})
\right],
\]
where \(\sigma(\cdot)\) is the sigmoid function. Matched queries are trained to reconstruct theoretical fragment representations, while the confidence head is trained to distinguish matched queries from unmatched ones.

The decoder is then fed with the confident imputed fragments and the observed peak representations:
\[
\widetilde{H}
=
\left[
h_{\mathrm{cls}};
\{\hat{u}_j:\sigma(\hat{r}_j)>\tau\};
h_1,\ldots,h_N
\right],
\]
where \(\tau\) is a confidence threshold. It then generates the peptide sequence autoregressively. The theoretical spectrum is required only to define training targets; at inference time, missing fragment representations are predicted directly from the observed spectrum.

We now describe our contributions. Figure \ref{fig:main_fig} provides an overview of all the contributions. Below, we describe each component in detail.

\subsection{Error-Guided Fragment Imputation Supervision}

The standard imputation objective assigns equal importance to all theoretical fragments. However, accurately reconstructing every missing fragment is not equally useful for peptide sequencing. In particular, fragments adjacent to amino acids that are already decoded correctly provide less useful training feedback than fragments associated with persistent decoder errors.

We therefore use the current decoder errors to adapt the representation-learning task in two complementary ways. First, we increase the imputation weight of theoretical fragments associated with difficult amino acids. Second, we construct an easy and a hard augmented spectrum view: the easy view exposes and strengthens difficult fragment evidence, whereas the hard view removes or weakens the same evidence to improve robustness. 

\subsubsection{Imputation Loss Re-Weighting.}

We first decode the peptide from the observed-spectrum branch and compute the cross-entropy loss at each valid amino acid position:
\[
\ell_{il}
=
-\log
P_\theta
\left(
y_{il}
\mid
y_{i,<l},\widetilde{H}_i,p_i
\right).
\]
Because raw cross-entropy values are unbounded, we map them to a bounded token-error score:
\[
e_{il}
=
1-\exp\left(-\ell_{il}\right).
\]

Each theoretical fragment \(k\) is annotated with its ion family and cleavage index. These annotations determine the amino acid positions adjacent to the corresponding cleavage.
Let \(\mathcal{N}_i(k)\) denote these positions together with a small local neighborhood. \footnote{In this work, we set the neighborhood size to 1.} The weight assigned to fragment \(k\) is
\[
w_{ik}
=
\operatorname{clip}
\left(
1+
\max_{l\in\mathcal{N}_i(k)}e_{il},
\,w_{\min},w_{\max}
\right),
\]
where $w_{\min} = 1.0$ and $w_{\max} = 2.0$ are hyperparameters. Thus, a theoretical fragment receives a larger weight when it is close to an amino acid that the decoder currently predicts poorly.

The decoder-conditioned imputation loss is
\[
\mathcal{L}_{\mathrm{imp}}
=
\frac{
\displaystyle
\sum_{(j,k)\in\pi_i}
w_{ik}
\|\hat{u}_{ij}-u_{ik}^{\star}\|_2^2
}{
\displaystyle
\sum_{(j,k)\in\pi_i}w_{ik}
}
+
\mathcal{L}_{\mathrm{conf}},
\]
where \(\pi_i\) is the bipartite matching for sample \(i\). The confidence loss \(\mathcal{L}_{\mathrm{conf}}\) is a binary cross-entropy objective in which matched queries are positive and unmatched queries are negative. The weights \(w_{ik}\) are also applied to matched positive queries, encouraging the confidence head to retain fragments associated with difficult residues.

\subsubsection{Observed Mass Spectrum Augmentations.}
Additionally, we use decoder errors to construct two peptide-conditioned augmented views of each observed spectrum. The conditioning signal is obtained from a teacher-forced forward pass through the observed-spectrum branch. For sample $i$, let $\ell_{il}$ denote the cross-entropy loss for ground-truth amino acid $l$, and let $L_i$ denote the peptide length, excluding padding and the end-of-sequence token. We define the sample-level decoding difficulty as
\[
\bar{\ell}_i = \frac{1}{L_i}\sum_{l=1}^{L_i}\ell_{il},
\qquad
\bar{e}_i = 1-\exp(-\bar{\ell}_i/\tau),
\]
and compute the augmentation scale
\[
s_i = \operatorname{clip}
\left(1+\alpha\bar{e}_i,\,s_{\min},\,s_{\max}\right),
\]
where $\tau=1$, $\alpha=1$, $s_{\min}=1$, and $s_{\max}=1.5$. This scale modulates a linear augmentation curriculum, with a lower base corruption strength for the easy view than for the hard view. Consequently, samples with larger decoding errors receive stronger background perturbations, subject to operator-specific upper bounds.

To localize the augmentation, we additionally convert amino-acid losses into residue-level errors,
$e_{il}=1-\exp(-\ell_{il}/\tau)$, and associate each theoretical fragment with the residues surrounding its cleavage site. The fragment weight $w_{ik}$ is determined by the largest residue-level error in this local neighborhood. These weights are normalized within each spectrum to obtain focus scores $f_{ik}\in[0,1]$, which control where the two views preserve or remove evidence. All conditioning quantities are detached from the computation graph.

\paragraph{View construction.}
Both views are constructed from the observed spectrum. We match each theoretical singly charged $b$- or $y$-ion to its nearest observed peak when their absolute $m/z$ difference is at most $0.5$ Da. This identifies matched observed peaks, unannotated observed peaks, and missing theoretical fragments. The perturbation design is informed by error analysis on a nine-species validation set separate from the test set, while the augmentation hyperparameters are selected by random search.

The easy view increases the retention probability and intensity of observed peaks associated with high decoder error, and increases the insertion probability of corresponding missing or dropped theoretical fragments. The hard view decreases these probabilities and attenuates the same observed peaks. Both views also incorporate stochastic peak removal, ion-family-specific dropout, contiguous fragmentation-ladder gaps, reduced visibility of modification-containing fragments, and insertion of structured noise peaks. Newly inserted theoretical peaks receive intensity and $m/z$ jitter. Thus, the easy view supplies additional evidence near difficult residues, whereas the hard view encourages reconstruction and decoding when that evidence is incomplete.

Both augmented views pass through the shared encoder, imputation module, and peptide decoder. Each receives imputation supervision from the theoretical fragment representations and sequence supervision from the ground-truth peptide; the easy and hard views contribute equally to the augmented-view losses. These augmentations are used only during training. Full construction details and hyperparameters are provided in Appendix~\ref{app:decoder-conditioned-augmentation}.

\subsection{Continuous Mass Rotary Attention}

The baseline spectrum encoder represents each peak independently using its absolute \(m/z\) and intensity. Thus, although the query and key vectors contain absolute mass information, vanilla self-attention does not explicitly expose the displacement between two peaks. Such displacements are informative for peptide fragmentation: Fragments derived from the same precursor exhibit residue-dependent \(m/z\) increments, and consecutive ions form approximately regular fragmentation ladders.

To introduce this inductive bias, we retain the baseline peak embeddings but replace vanilla spectrum self-attention with a continuous-\(m/z\) adaptation of Rotary Position Embedding (RoPE) \cite{SU2024127063}. Unlike conventional RoPE, whose rotation angles are determined by discrete token indices, our rotation angles are determined directly by the continuous \(m/z\) coordinate of each peak.

Let \(d_h\) be the dimension of one attention head and assume it is even. We partition each query and key vector into \(d_h/2\) two-dimensional pairs. For pair \(r\), we use the fixed wavelength
\[
\lambda_r
=
\lambda_{\min}
\left(
\frac{\lambda_{\max}}{\lambda_{\min}}
\right)^{\frac{r}{d_h/2-1}},
\qquad
r=0,\ldots,\frac{d_h}{2}-1,
\]
so that the wavelengths are logarithmically spaced between \(\lambda_{\min}\) = 1 and \(\lambda_{\max}\) = 10,000. A peak at \(m/z\) value \(m\)
induces the phase
\[
\phi_r(m)=\frac{2\pi m}{\lambda_r}
\]
and the corresponding rotation
\[
R_r(m)
=
\begin{bmatrix}
\cos\phi_r(m) & -\sin\phi_r(m)\\
\sin\phi_r(m) & \cos\phi_r(m)
\end{bmatrix}.
\]
Let \(R(m)=\operatorname{diag}(R_0(m),\ldots,R_{d_h/2-1}(m))\). We rotate only the query and key vectors, leaving the value vectors unchanged:
\[
\widetilde q_i=R(m_i)q_i,
\qquad
\widetilde k_j=R(m_j)k_j.
\]
The resulting attention logit is
\[
\begin{aligned}
a_{ij}
&=
\frac{\widetilde q_i^{\top}\widetilde k_j}{\sqrt{d_h}}\\
&=
\frac{q_i^{\top}R(m_i)^{\top}R(m_j)k_j}{\sqrt{d_h}}\\
&=
\frac{q_i^{\top}R(m_j-m_i)k_j}{\sqrt{d_h}}.
\end{aligned}
\]
The final equality follows from the composition property of planar rotations. Consequently, the rotary component of the attention score depends on the relative displacement \(\Delta m_{ij}=m_j-m_i\), rather than on the two coordinates only independently. The same learned query-key interaction can therefore recognize a characteristic fragmentation gap regardless of where it occurs in the spectrum.

\subsection{Training Objective and Inference}

The complete objective combines peptide decoding from the observed spectrum, peptide decoding from the theoretical spectrum, imputation on the observed spectrum, and imputation and decoding on the augmented views:
\begin{equation}
\begin{aligned}
\mathcal{L}
&=
\mathcal{L}_{\mathrm{dec}}(x)
+
\mathcal{L}_{\mathrm{dec}}(x^\star)
+
\mathcal{L}_{\mathrm{imp}}(x)
\\
&\quad+
\mathcal{L}_{\mathrm{imp}}(\widetilde{x})
+
\mathcal{L}_{\mathrm{dec}}(\widetilde{x}),
\end{aligned}
\end{equation}
where \(x^\star\) is the theoretical spectrum and \(\widetilde{x}\) denotes the set of training-time easy and hard augmented views. The augmented-view losses are computed as weighted averages over the two views.

At inference time, the model requires only the observed spectrum and precursor metadata. The theoretical spectrum, decoder-conditioned weights, and augmented views are discarded. The observed spectrum is encoded using continuous mass (m/z) rotary attention, the learned imputation queries predict candidate missing-fragment representations, and low-confidence candidates are removed using the threshold \(\tau\) (we set it to 0.8 following \citet{du2025latent}). Finally, the peptide decoder attends jointly to the observed and imputed spectrum representations and generates the peptide sequence using beam search.

\section{Experiments}

\subsection{Datasets}

Consistent with the benchmark established by \citet{zhou2024novobench}, we evaluate our method on three datasets: Nine-species and Seven-species \cite{doi:10.1073/pnas.1705691114}, and HC-PT \cite{Eloff2025}. The Nine-species dataset, which has been widely adopted in prior research, includes mass spectra collected from nine species. Spectra from yeast are reserved for testing, while data from the remaining eight species are used for model training and validation. Similarly, the Seven-species dataset contains mass spectra from seven species, with yeast serving as the held-out test species and the other six used for training and validation. The HC-PT dataset comprises spectra from human-derived peptides, including synthetic tryptic peptides covering canonical human proteins and their isoforms, peptides produced using alternative proteases, and human leukocyte antigen–associated peptides.

\subsection{Baselines}

We compare our proposed \methodname against strong baselines, including DeepNovo \cite{doi:10.1073/pnas.1705691114}, PointNovo \cite{10.1093/bioadv/vbad057}, CasaNovo \cite{Yilmaz2024}, AdaNovo \cite{xia2024adanovo}, LIPNovo \cite{du2025latent}, and the recent concurrent work LIPNovo+ \cite{du2026lipnovo}. Unless otherwise noted, the reported results are taken from NovoBench. Because LIPNovo is our most direct competitor, we report both its NovoBench results and our reproduced results to ensure a fair comparison. We refer to the reproduced version as the ``baseline'' throughout the paper and in the result tables.

\subsection{Evaluation Metrics}

We assess \textit{de novo} peptide sequencing at three levels of granularity: amino acid, PTM, and complete peptide. An amino acid prediction is considered correct if its mass differs from that of the corresponding reference residue by less than $0.1$~Da and its prefix or suffix mass differs from that of the reference sequence by less than $0.5$~Da \cite{doi:10.1073/pnas.1705691114}. We report precision and recall at both the amino acid and PTM levels.

For peptide-level evaluation, a prediction is regarded as correct only when the entire predicted sequence exactly matches the reference sequence. We report the peptide-level precision.  Finally, we report the area under the precision–recall curve (AUC) to characterize performance across confidence-score thresholds.

\subsection{Implementation Details}

To maintain a controlled comparison with prior work, we adopt the hyperparameter configuration reported by \citet{du2025latent}. Our model follows the approach of \citet{du2025latent}, which builds on an implementation of Casanovo \cite{Yilmaz2024} by incorporating complementary spectra \cite{10.1093/bib/bbae021} as an additional input modality and introducing an imputation module.

The encoder and decoder each contain nine Transformer layers, with a hidden dimension of 512, a feed-forward dimension of 1024, and eight attention heads. The imputation module consists of three layers and uses 100 peak queries. Input spectra are truncated or padded to a maximum of 150 peaks, and the maximum peptide sequence length is set to 100 residues. The decoder vocabulary contains tokens for the 20 standard amino acids, three post-translational modifications, and an end-of-sequence token.

We train once the model using distributed data parallelism on four NVIDIA A100 GPUs with a global batch size of 32. Training is performed for at most 30 epochs using a learning rate of $5 \times 10^{-4}$, with early stopping applied when the validation performance does not improve for five consecutive epochs. During inference, peptide sequences are generated using beam search with a beam width of five.

\section{Results and Analysis}

\subsection{Main Results}
\begin{table*}[!htbp]
\centering
\small
\setlength{\tabcolsep}{4.5pt}
\renewcommand{\arraystretch}{1.12}

\begin{adjustbox}{max width=\textwidth}
\begin{tabular}{@{}l c *{12}{c}@{}}
\toprule
\multirow{3}{*}{\textbf{Method}}
& \multirow{3}{*}{\textbf{Year}}
& \multicolumn{6}{c}{\textbf{Amino Acid Level Performance}}
& \multicolumn{6}{c}{\textbf{Peptide Level Performance}} \\
\cmidrule(lr){3-8}
\cmidrule(lr){9-14}

&
& \multicolumn{2}{c}{\textbf{Nine-Species}}
& \multicolumn{2}{c}{\textbf{Seven-Species}}
& \multicolumn{2}{c}{\textbf{HC-PT}}
& \multicolumn{2}{c}{\textbf{Nine-Species}}
& \multicolumn{2}{c}{\textbf{Seven-Species}}
& \multicolumn{2}{c}{\textbf{HC-PT}} \\
\cmidrule(lr){3-4}
\cmidrule(lr){5-6}
\cmidrule(lr){7-8}
\cmidrule(lr){9-10}
\cmidrule(lr){11-12}
\cmidrule(lr){13-14}

&
& Prec. & Rec.
& Prec. & Rec.
& Prec. & Rec.
& Prec. & AUC
& Prec. & AUC
& Prec. & AUC \\
\midrule

PEAKS \cite{https://doi.org/10.1002/rcm.1196}
& 2003
& 0.748 & --
& -- & --
& -- & --
& 0.428 & --
& -- & --
& -- & -- \\

DeepNovo \cite{doi:10.1073/pnas.1705691114}
& 2017
& 0.696 & 0.638
& 0.492 & 0.433
& 0.531 & 0.534
& 0.428 & 0.376
& 0.204 & 0.136
& 0.313 & 0.255 \\

PointNovo \cite{Qiao2021}
& 2021
& 0.740 & 0.671
& 0.196 & 0.169
& 0.623 & 0.622
& 0.480 & 0.436
& 0.022 & 0.007
& 0.419 & 0.373 \\

CasaNovo \cite{Yilmaz2024}
& 2024
& 0.697 & 0.696
& 0.322 & 0.327
& 0.442 & 0.453
& 0.481 & 0.439
& 0.119 & 0.084
& 0.211 & 0.177 \\

AdaNovo \cite{xia2024adanovo}
& 2024
& 0.698 & 0.709
& 0.379 & 0.385
& 0.442 & 0.451
& 0.505 & 0.469
& 0.174 & 0.135
& 0.212 & 0.178 \\

InstaNovo \cite{Eloff2025}
& 2025
& 0.420 & 0.395
& 0.192 & 0.176
& 0.289 & 0.285
& 0.164 & 0.123
& 0.031 & 0.009
& 0.057 & 0.034 \\

LIPNovo \cite{du2025latent}
& 2025
& 0.797 & 0.797
& 0.557 & 0.560
& 0.637 & 0.643
& 0.582 & 0.547
& 0.327 & 0.281
& 0.458 & 0.427 \\

LIPNovo+ \cite{du2026lipnovo}
& 2026
& 0.831 & 0.831
& 0.619 & 0.622
& 0.698 & 0.694
& 0.637 & 0.612
& 0.383 & 0.337
& 0.519 & 0.490 \\

\midrule

\rowcolor{gray!8}
Baseline
& 2025
& 0.794 & 0.796
& 0.553 & 0.554
& 0.653 & 0.654
& 0.583 & 0.549
& 0.325 & 0.270
& 0.469 & 0.438 \\

\rowcolor{gray!15}
\textbf{\methodname (Ours)}
& \textbf{2026}
& \textbf{0.848} & \textbf{0.848}
& \textbf{0.632} & \textbf{0.635}
& \textbf{0.704} & \textbf{0.702}
& \textbf{0.674} & \textbf{0.646}
& \textbf{0.421} & \textbf{0.378}
& \textbf{0.532} & \textbf{0.502} \\

\bottomrule
\end{tabular}
\end{adjustbox}

\caption{Amino-acid- and peptide-level performance on the Nine-Species, Seven-Species, and HC-PT datasets. Best results are shown in bold.}
\label{tab:performance-comparison}
\end{table*}

The empirical results of our experiments are summarized in Tables \ref{tab:performance-comparison} and \ref{tab:ptm-level-performance}. Overall, \methodname consistently outperforms the baseline methods across all datasets and evaluation metrics. Table \ref{tab:performance-comparison} presents the amino acid- and peptide-level results. At the amino acid level, \methodname achieves the highest precision and recall on all datasets and exceeds the strongest baseline (LIPNovo) by an average of 6.4 percentage points in precision. Similarly, at the peptide level, \methodname outperforms all baseline methods across all metrics and surpasses the baseline by an average of 8.7 percentage points in precision. These improvements are consistent across the three evaluated datasets. Table \ref{tab:ptm-level-performance} presents the PTM-level results and shows that \methodname outperforms the strongest baseline by an average of 5.6 percentage points in precision. 

In addition to the preceding comparisons, we compare \methodname with the recently released, \textit{concurrent} LIPNovo+ \cite{du2026lipnovo}. As shown in Tables \ref{tab:performance-comparison} and \ref{tab:ptm-level-performance}, \methodname consistently outperforms LIPNovo+ across all metrics on all three datasets. In particular, \methodname improves amino-acid-level, peptide-level, and PTM-level precision by 1.20, 2.93, and 2.10 average percentage points, respectively.


\begin{table}[t]
\centering
\small
\setlength{\tabcolsep}{8pt}
\renewcommand{\arraystretch}{1.15}

\begin{adjustbox}{max width=\textwidth}
\resizebox{0.48\textwidth}{!}{%
\begin{tabular}{@{}l *{6}{c}@{}}
\toprule

\multirow{2}{*}{\textbf{Method}}
& \multicolumn{2}{c}{\textbf{Nine-Species}}
& \multicolumn{2}{c}{\textbf{Seven-Species}}
& \multicolumn{2}{c}{\textbf{HC-PT}} \\[-1pt]

\cmidrule(lr){2-3}
\cmidrule(lr){4-5}
\cmidrule(lr){6-7}

& Precision & Recall
& Precision & Recall
& Precision & Recall \\

\midrule

DeepNovo
& 0.576 & 0.529
& 0.391 & 0.373
& 0.626 & 0.615 \\

PointNovo
& 0.629 & 0.546
& 0.117 & 0.094
& 0.676 & 0.740 \\

InstaNovo
& 0.443 & 0.294
& 0.126 & 0.115
& 0.350 & 0.261 \\

CasaNovo
& 0.706 & 0.566
& 0.360 & 0.251
& 0.501 & 0.460 \\

AdaNovo
& 0.652 & 0.570
& 0.448 & 0.321
& 0.552 & 0.482 \\

LIPNovo
& 0.765 & 0.656
& 0.604 & 0.498
& 0.732 & 0.745 \\

LIPNovo+
& 0.801 & 0.678
& 0.648 & 0.561
& 0.757 & 0.805 \\

\midrule

\rowcolor{gray!8}
Baseline
& 0.773 & 0.646
& 0.602 & 0.500
& 0.690 & 0.762 \\

\rowcolor{gray!15}
\textbf{\methodname (Ours)}
& \textbf{0.815} & \textbf{0.731}
& \textbf{0.665} & \textbf{0.565}
& \textbf{0.790} & \textbf{0.817} \\

\bottomrule
\end{tabular}}
\end{adjustbox}

\caption{PTM-level performance on the Nine-Species, Seven-Species, and HC-PT datasets. Best results are shown in bold.}
\label{tab:ptm-level-performance}
\end{table}

\subsection{Ablation Study and Analysis}

\subsubsection{Component Ablation.}

Table~\ref{tab:ablation} presents a cumulative ablation study of \methodname on the nine-species dataset, where each component is added on top of all components included in the preceding rows. Imputation-loss reweighting provides only marginal changes, improving amino-acid precision and recall by $0.4$ and $0.2$ percentage points, respectively, and peptide precision by $0.3$ percentage points. Adding augmented views yields the largest incremental gains, improving amino-acid precision and recall by $3.1$ percentage points each, and peptide precision by $5.3$ percentage points. Mass Rotary Attention further improves these metrics by $1.9$, $1.9$, and $3.5$ percentage points, respectively. Overall, the complete model improves upon the LIPNovo baseline by $5.4$ and $5.2$ percentage points in amino-acid precision and recall, respectively, and by $9.1$ percentage points in peptide precision.

\subsubsection{Mechanistic Analysis of Mass Rotary Attention.}

To assess whether Mass Rotary Attention improves access to chemically related peaks, we formulate spectrum self-attention as a neighbor-retrieval task. For each spectrum, we match observed peaks to theoretical fragments of the ground-truth peptide, and label pairs corresponding to adjacent \(b\)-ions or \(y\)-ions as true ladder neighbors. For every eligible query peak, encoder layer, and attention head, we rank the remaining observed peaks using the query–key attention logits and compute Hit@\(K\), indicating whether at least one true ladder neighbors appear among the \(K\) highest-ranked peaks. We average scores across queries, heads, and layers within each spectrum and then macro-average across spectra. We compare the vanilla encoder with Mass Rotary Attention and include two controls: a within-spectrum intervention that permutes only the mass assignments supplied to the rotary operation while preserving the peak embeddings and intensities, and an exact random-ranking baseline accounting for the number of candidate and positive peaks. 

\begin{figure}[h]
    \centering
    \includegraphics[width=0.49\textwidth]{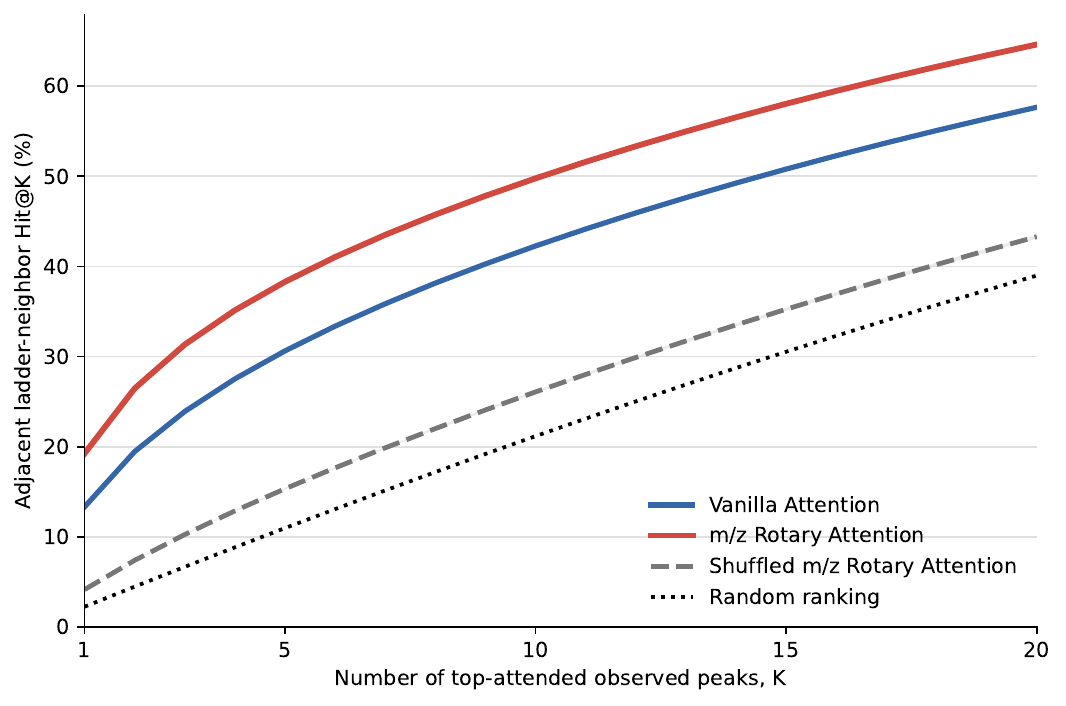}
    \caption{Adjacent ion-ladder neighbor retrieval performance of vanilla and mass-aware rotary attention on Nine-Species dataset.}
    \label{fig:attention-ladder-analysis}
\end{figure}

As shown in Figure \ref{fig:attention-ladder-analysis}, Mass Rotary Attention consistently improves adjacent ion-ladder neighbor retrieval across all values of \(K\). In particular, Hit@1 increases from approximately 13\% with vanilla attention to 19\% with Mass Rotary Attention, while Hit@20 improves from approximately 58\% to 65\%. Shuffling the mass assignments used by the rotary operation substantially reduces retrieval performance, bringing it closer to random ranking and below the vanilla model. These results show that the improvement arises from correctly encoding relative mass relationships, enabling self-attention to prioritize peptide-consistent peak neighbors.

\begin{table}[t]
    \centering
    \small
    \setlength{\tabcolsep}{4.5pt}
    \resizebox{0.48\textwidth}{!}{%
    \begin{tabular}{lcccc}
        \toprule
        \multirow{2}{*}{\textbf{Configuration}}
        & \multicolumn{2}{c}{\textbf{Amino Acid Level}}
        & \multicolumn{2}{c}{\textbf{Peptide Level}} \\
        \cmidrule(lr){2-3} \cmidrule(lr){4-5}
        & Precision
        & Recall
        & Precision
        & AUC \\
        \midrule
        Baseline
        & 0.794 & 0.796 & 0.583 & 0.550 \\
        $+$ Imputation-loss reweighting
        & 0.798 & 0.798 & 0.586 & 0.549 \\
        $+$ Augmented Views
        & 0.829
        & 0.829
        & 0.639
        & 0.607 \\
        $+$ Mass Rotary Attention
        & \textbf{0.848}
        & \textbf{0.848}
        & \textbf{0.674}
        & \textbf{0.646} \\
        \bottomrule
    \end{tabular}}
    \caption{Cumulative ablation study of \methodname on the Nine-Species dataset. Components are added sequentially to the reproduced baseline. Best results are shown in bold.}
    \label{tab:ablation}
\end{table}
\begin{table}[t]
    \centering
    
    \small
    \setlength{\tabcolsep}{3.5pt}
    \renewcommand{\arraystretch}{0.95}
    \begin{tabular}{lcccc}
        \toprule
        & \multicolumn{2}{c}{Training}
        & \multicolumn{2}{c}{Inference} \\
        \cmidrule(lr){2-3}\cmidrule(lr){4-5}
        \textbf{Model}
        & \textbf{Thrpt.}$\uparrow$
        & \textbf{Mem.}$\downarrow$
        & \textbf{Thrpt.}$\uparrow$
        & \textbf{Mem.}$\downarrow$ \\
        \midrule
        Baseline
        & 1.00$\times$ & 1.00$\times$
        & 1.00$\times$ & 1.00$\times$ \\
        \methodname-Pairwise
        & 0.27$\times$ & 4.02$\times$
        & 1.00$\times$ & 3.86$\times$ \\
        \textbf{\methodname (ours)}
        & 0.30$\times$ & 2.08$\times$
        & 1.00$\times$ & 1.01$\times$ \\
        \bottomrule
    \end{tabular}
    \caption{Training and inference efficiency of the baseline, \methodname-Pairwise, and our \methodname. Throughput and peak memory usage are normalized to the baseline; higher throughput and lower memory usage are better.}
    \label{tab:rope-pairwise-comparison}
\end{table}

\subsubsection{Time and Memory Overhead.}

We compare the computational overhead of \methodname and LIPNovo in terms of throughput and peak memory usage. We also include \methodname-Pairwise in this comparison, a variant that replaces rotary mass embeddings with an additive attention bias computed from a precomputed pairwise mass-difference matrix, following \citet{10.1021/acs.jproteome.5c00063}. This variant allows us to directly evaluate whether Mass Rotary Attention represents mass relationships more efficiently than explicitly constructing and storing a pairwise mass-difference matrix. Table~\ref{tab:rope-pairwise-comparison} summarizes the results relative to LIPNovo. During training, \methodname achieves $(0.30\times)$ the baseline throughput while using $(2.08\times)$ the baseline memory, compared with $(0.27\times)$ throughput and $(4.02\times)$ memory for \methodname-Pairwise. At inference, both variants maintain baseline throughput; however, \methodname uses only $(1.01\times)$ the baseline memory, whereas \methodname-Pairwise requires $(3.86\times)$. Finally, these results demonstrate that rotary mass embeddings capture mass relationships substantially more memory-efficiently than a precomputed pairwise mass-difference matrix, particularly during inference.

Both LIPNovo and GyroNovo converged at epoch 13 under the same early-stopping protocol, despite a maximum training budget of 30 epochs. Therefore, the observed improvements are not attributable to GyroNovo receiving more optimization epochs, although its multi-view training still entails greater computation per epoch.

\section{Conclusion}

In this paper, we propose \methodname, which addresses two major limitations of existing Transformer-based \textit{de novo} peptide sequencing methods: uniformly supervising the imputation of missing fragment ions and insufficiently modeling relative mass relationships between spectral peaks. First, our decoder-guided learning strategy uses current amino-acid prediction difficulty to prioritize fragments that are most relevant to unresolved residues. The same signal guides complementary easy and hard spectrum augmentations, providing clearer evidence for difficult residues while improving robustness to incomplete fragmentation ladders. Second, continuous mass (m/z) rotary attention explicitly incorporates relative mass differences into spectrum self-attention without introducing additional learned positional parameters.

These components retain the latent-imputation framework and require no theoretical spectrum or augmentation procedure at inference time. Our results demonstrate that aligning fragment representation learning with decoder uncertainty, together with an appropriate relative-mass inductive bias, provides a more effective approach to learning from sparse and noisy tandem mass spectra. We hope this work inspires further investigation into task-aware spectrum representation learning and physically informed attention mechanisms for database-free peptide sequencing.



\section*{Acknowledgments}
This work was supported by the UBC AI and Health Network at the University of British Columbia. Muhammad Abdul-Mageed also acknowledges support from Canada Research Chairs (CRC), Canada Foundation for Innovation (CFI; 37771), Digital Research Alliance of Canada,\footnote{\url{https://alliancecan.ca}} and UBC ARC-Sockeye.

\bibliography{aaai2027}

\clearpage
\appendix
\setcounter{secnumdepth}{2}

\section{Decoder-Conditioned Spectrum Augmentation}
\label{app:decoder-conditioned-augmentation}

\paragraph{Decoder-derived conditioning.}
Conditioning uses the current decoder's teacher-forced predictions on the observed-spectrum branch. The token losses used for conditioning are ordinary cross-entropy losses without label smoothing; padding and end-of-sequence tokens are excluded. The sample-level scale $s_i$ controls overall corruption strength, while fragment-level weights determine its spatial distribution along the fragmentation ladders.

For theoretical fragment $k$, let $\mathcal{N}_{ik}$ contain the two residues adjacent to its cleavage site and their immediate neighbors, restricted to valid peptide positions. This corresponds to a neighborhood radius of one around each cleavage-adjacent residue. The mapping accounts for the ion family and the decoder's sequence direction. We define
\[
w_{ik}
=
\operatorname{clip}
\left(
1+\beta\max_{l\in\mathcal{N}_{ik}}e_{il},
\,w_{\min},\,w_{\max}
\right),
\]
where $\beta=1$, $w_{\min}=1$, and $w_{\max}=2$. Taking the maximum ensures that a single difficult residue can emphasize nearby fragments without being diluted by easier neighboring residues.

The augmentation focus is
\[
f_{ik}
=
\begin{cases}
\dfrac{w_{ik}-1}{\max_{k'}(w_{ik'}-1)},
& \max_{k'}(w_{ik'}-1)>10^{-8},\\[6pt]
0, & \text{otherwise}.
\end{cases}
\]
Accordingly, the most emphasized fragment in a spectrum has focus one whenever the conditioning signal is nonzero. This normalization makes focus a relative measure within each spectrum; the sample scale $s_i$ separately captures absolute decoding difficulty. Decoder-derived weights are computed for the theoretical targets retained by the imputation branch, which contains at most 100 targets in our configuration. Additional theoretical fragments receive the neutral weight $w_{ik}=1$.

\paragraph{Curriculum and view-specific strength.}
Let $t\in[0,1]$ denote training progress, computed as the zero-based current epoch divided by the configured maximum number of epochs. We represent corruption strength by
\[
\mathbf{d}
=
(d_{\mathrm{miss}},d_b,d_y,d_{\mathrm{gap}},
 \sigma_I,\sigma_m,\lambda_N,d_{\mathrm{PTM}}),
\]
corresponding to general peak missingness, $b$-ion dropout, $y$-ion dropout, ladder-gap probability, intensity jitter, $m/z$ jitter, structured-noise count, and PTM visibility reduction. The curriculum interpolates linearly from the zero vector to
\[
\mathbf{d}_{\max}
=
(0.5,\,0.5,\,0.5,\,0.8,\,0.5,\,0.2,\,24,\,0.8).
\]
For view $v$, the effective strength is
\[
\mathbf{d}_{i}^{(v)}
=
\operatorname{clip}_{[0,\mathbf{d}_{\max}]}
\left(
s_i\,a_v\,t\,
\bigl(\mathbf{m}_i\odot\mathbf{d}_{\max}\bigr)
\right),
\]
where $a_{\mathrm{easy}}=0.5$, $a_{\mathrm{hard}}=1$, and clipping is componentwise. The multiplier $\mathbf{m}_i$ equals one except for its PTM component, which is $1.2$ for peptides containing modified residues. These entries are operator strengths: depending on the operator, they act as probability penalties, Gaussian standard deviations, or Poisson count parameters.

The fixed retention and insertion probabilities and decoder-focus adjustments remain active even when $t=0$. The curriculum therefore increases additional corruption over training rather than delaying all augmentation until later epochs.

\paragraph{Matching and observed-peak retention.}
Each theoretical fragment is independently assigned to its nearest observed peak if their absolute $m/z$ difference is at most $0.5$; matching is not constrained to be one-to-one. When multiple fragments match the same observed peak, the peak receives their maximum focus score. Unannotated observed peaks receive focus zero.

For observed peak $j$, let $F_{ij}$ denote its focus, and let $B_{ij}$, $Y_{ij}$, and $P_{ij}$ indicate its assigned $b$-ion, $y$-ion, and modification-containing annotations. These indicators are zero for unannotated peaks. Before applying ladder gaps, the retention probability is

\begin{align*}
p_{ij}^{\mathrm{keep},v} = \operatorname{clip}_{[0,1]} \Bigl[
  & p_{ij}^{0} - 0.20 d_{\mathrm{miss}} \\
  & - 0.20 B_{ij} d_b - 0.20 Y_{ij} d_y \\
  & - 0.10 P_{ij} d_{\mathrm{PTM}} + \eta_v \, 0.40 F_{ij} \Bigr],
\end{align*}

where $p_{ij}^{0}=0.90$ for matched peaks and $0.70$ for unannotated peaks, and $\eta_{\mathrm{easy}}=+1$ and $\eta_{\mathrm{hard}}=-1$. Here and below, operator strengths are those of the corresponding sample and view.

Retained observed peaks preserve their measured $m/z$ values. Their intensities are adjusted as
\[
I_{ij}^{(v)}
=
I_{ij}\left(1+\eta_v\,0.20F_{ij}\right).
\]
Thus, maximally focused observed peaks are strengthened by $20\%$ in the easy view and attenuated by $20\%$ in the hard view. Unannotated observed peaks retain their original intensities. No additional independent focus-based forced-drop stage is used.

\paragraph{Insertion of missing or dropped fragments.}
A theoretical fragment is eligible for insertion if it has no observed match or its matched observed peak was removed. Its insertion probability is

\begin{align*}
p_{ik}^{\mathrm{add},v} = \operatorname{clip}_{[0,0.90]} \Bigl[
  & 0.30 \\
  & + \eta_v \, 0.30 f_{ik} - 0.15 d_{\mathrm{miss}} \\
  & - 0.15 P_{ik} d_{\mathrm{PTM}} \Bigr],
\end{align*}

subject to the ladder-gap restrictions below. The insertion operation can therefore restore both originally missing evidence and evidence removed earlier in view construction.

For a fragment of ordinal index $r\in\{1,\ldots,L_i-1\}$, the baseline synthetic intensity is
\begin{align*}
\widehat{I}_{ik} &= R_i \biggl[ 0.75 + 0.35 \biggl( 1 - 2 \, \biggl| \frac{r}{L_i - 1} - \frac{1}{2} \biggr| \biggr) \biggr], \\
&\quad\text{where } R_i = \max \Bigl( 1, \, \max_j I_{ij} \Bigr).
\end{align*}
The same baseline profile is used for $b$- and $y$-ions, with no additional PTM-specific intensity boost. Inserted fragment intensities and locations are perturbed according to
\[
I_{ik}^{\mathrm{add}}
=
\widehat I_{ik}\max(0.05,1+\epsilon_I),
\qquad
m_{ik}^{\mathrm{add}}
=
m_{ik}^{\mathrm{theory}}+\epsilon_m,
\]
where $\epsilon_I\sim\mathcal{N}(0,\sigma_I^2)$ and
$\epsilon_m\sim\mathcal{N}(0,\sigma_m^2)$.

\paragraph{Contiguous ladder gaps.}
A ladder-gap event is sampled with probability $d_{\mathrm{gap}}$. When activated, its run length is sampled uniformly from
\[
\left\{1,\ldots,
\min\left(L_i-1,\,
1+\operatorname{round}(3d_{\mathrm{gap}})\right)
\right\},
\]
and a valid starting index is chosen uniformly. Under the configured maximum $d_{\mathrm{gap}}=0.8$, this produces runs of at most three consecutive fragment indices. The selected ordinal indices are applied to both ion ladders, grouping $b_r$ and $y_r$ by fragment index.

In the hard view, observed peaks assigned to the selected indices are removed and theoretical insertion at those indices is disabled. In the easy view, matched peaks with focus at least $0.5$ are exempt from this additional ladder-gap removal, although they remain subject to the ordinary retention draw. Insertion within an easy-view gap is permitted only for fragments with focus strictly greater than $0.5$.

\paragraph{Structured noise and final spectrum assembly.}
In addition to retaining a subset of unannotated observed peaks, we sample
\[
N_i^{(v)}\sim\operatorname{Poisson}(0.35\lambda_N)
\]
candidate structured-noise peaks. Each candidate is generated from a sampled theoretical parent fragment by subtracting an offset of either $18.010565$ or $17.026549$, corresponding to water- or ammonia-loss masses, and adding Gaussian $m/z$ jitter with standard deviation $0.01$. Its intensity is sampled uniformly between $0.08$ and $0.35$ times the parent's baseline synthetic intensity and capped at $0.35R_i$. Noise candidates below $m/z=50$ are discarded.

Retained observed peaks, inserted theoretical fragments, and structured-noise peaks are combined. Invalid or nonpositive entries are removed, the 150 most intense peaks are retained if necessary, and the resulting spectrum is sorted by $m/z$. If no peak survives, the strongest baseline theoretical fragment is retained to ensure a nonempty input.

\paragraph{Supervision and hyperparameter selection.}
Both views retain the original precursor information and ground-truth peptide label. They are processed by the same encoder, imputation module, and decoder as the observed-spectrum branch. The fragment weights $w_{ik}$ also emphasize difficult targets in the imputation objective. Easy- and hard-view imputation losses are averaged with equal weights, as are their sequence losses. Conditioning signals are recomputed from the current observed-spectrum predictions during training and are treated as constants when constructing the augmented views.

The augmentation hyperparameters reported above were selected by random search. Error analysis on the nine-species validation set, held separate from the test set, motivated the perturbation types; their numerical strengths and probability coefficients are tunable augmentation parameters.


\end{document}